\documentclass[sigconf]{acmart}
\usepackage{subfigure}
\usepackage{multirow}
\usepackage{hyperref}
\usepackage{makecell}
\usepackage{soul}
\usepackage{adjustbox}

\AtBeginDocument{%
  \providecommand\BibTeX{{%
    \normalfont B\kern-0.5em{\scshape i\kern-0.25em b}\kern-0.8em\TeX}}}

\setcopyright{none}
\renewcommand\footnotetextcopyrightpermission[1]{} % removes footnote with conference information in first column
\begin{document}

\title{Model Debiasing via Gradient-based Explanation on Representation}

\author{Jindi Zhang}
\email{jd.zhang@my.cityu.edu.hk}
\affiliation{%
  \institution{City University of Hong Kong}
  \country{Hong Kong SAR}
}

\author{Luning Wang}
\email{wangluning2@huawei.com}
\affiliation{%
  \institution{Hong Kong Research Center, Huawei}
  \country{Hong Kong SAR}
}

\author{Dan Su}
\email{dasu@nvidia.com}
\affiliation{%
  \institution{NVIDIA Research}
  \country{Hong Kong SAR}
}

\author{Yongxiang Huang}
\email{huang.yongxiang2@huawei.com}
\affiliation{%
  \institution{Hong Kong Research Center, Huawei}
  \country{Hong Kong SAR}
}

\author{Caleb Chen Cao}
\email{goupcaleb@gmail.com}
\affiliation{%
  \institution{The Hong Kong University of Science and Technology}
  \country{Hong Kong SAR}
}

\author{Lei Chen}
\email{leichen@cse.ust.hk}
\affiliation{%
  \institution{The Hong Kong University of Science and Technology}
  \country{Hong Kong SAR}
}

\renewcommand{\shortauthors}{Zhang, et al.}

\begin{abstract}
Machine learning systems produce biased results towards certain demographic groups, known as the fairness problem. Recent approaches to tackle this problem learn a latent code (i.e., representation) through disentangled representation learning and then discard the latent code dimensions correlated with sensitive attributes (e.g., gender). Nevertheless, these approaches may suffer from incomplete disentanglement and overlook proxy attributes (proxies for sensitive attributes) when processing real-world data, especially for unstructured data, causing performance degradation in fairness and loss of useful information for downstream tasks. In this paper, we propose a novel fairness framework that performs debiasing with regard to both sensitive attributes and proxy attributes, which boosts the prediction performance of downstream task models without complete disentanglement. The main idea is to, first, leverage gradient-based explanation to find two model focuses, 1) one focus for predicting sensitive attributes and 2) the other focus for predicting downstream task labels, and second, use them to perturb the latent code that guides the training of downstream task models towards fairness and utility goals. We show empirically that our framework works with both disentangled and non-disentangled representation learning methods and achieves better fairness-accuracy trade-off on unstructured and structured datasets than previous state-of-the-art approaches.
\end{abstract}

%    <concept>
%        <concept_id>10010147.10010257.10010293</concept_id>
%        <concept_desc>Computing methodologies~Machine learning approaches</concept_desc>
%        <concept_significance>500</concept_significance>
%        </concept>
%  </ccs2012>

% \ccsdesc[500]{Computing methodologies~Machine learning approaches}




\maketitle
\pagestyle{plain}

\section{Introduction}
\label{introduction}

% Machine learning systems have been widely adopted in the real-world applications due to their powerful inference and generalization capability. However, 

Machine learning systems are reported to generate preferential predictions for some demographic groups and prejudiced predictions for others in many high-stake fields, such as loan offers, exam grading, school admission, and parole approval~\cite{calders2010three, smith2020algorithmic, marcinkowski2020implications, dressel2018accuracy}. This is known as the fairness problem in machine learning. Such fairness problems may cause long-term and high impacts on the life of vulnerable groups~\cite{d20fairness}.

% Such fairness problems cause a long-term negative impact on vulnerable groups~\cite{d20fairness}, and is receiving more and more attentions from the public and policy makers~\cite{voigt17eu,jillson21aiming}. Thus, debiasing machine learning systems is more necessary than ever before.

To tackle the fairness problem, early studies use adversarial training and regularization to force the model not to pay attention to sensitive information during training~\cite{ganin16domain,pfohl19creating,kim19learning,wang21understanding,zhang21omnifair,zhang18mitigating}. And other works focus on learning fair (debiased) representations for downstream tasks~\cite{park21learning,yan21equitensors,lahoti19ifair,lahoti19operationalizing,locatello19fairness,sweeney20reducing}. These methods usually specify the task attributes or the sensitive attributes before training, resulting in inflexibility~\cite{creager19flexibly}.

To increase flexibility, the up-to-date approaches are to leverage disentangled representation learning methods to learn the disentangled latent code in which every dimension only contains one factor of variation and is optimized to be independent of each other~\cite{higgins16beta,kim18disentangling,chen18isolating,creager19flexibly}, and then remove the dimensions correlated with sensitive attributes before using the code to train downstream task models~\cite{park21learning,creager19flexibly}.

However, because it is extremely difficult to enumerate all the factors of variation in real-world data~\cite{locatello19challenging}, the number of the latent code dimensions is usually smaller than the real number of factors of variation. This results in incomplete disentanglement in the latent code, which poses two major challenges when we process real-world data, in particular, unstructured data such as images, with debiasing methods based on disentangled representation learning.
% \vspace{-5pt}
\begin{itemize}
\item
First, it is challenging to avoid information loss for downstream tasks during the debiasing process. Since the latent code is usually incompletely disentangled, critical information for downstream tasks can be lost when we remove the dimensions correlated with sensitive attributes from the latent code, causing degradation in prediction accuracy.
\item
Second, it is challenging to cover all sensitive information in proxy attributes\footnote{For example, when the sensitive attribute is gender, corresponding proxy attributes can be hair length, beard, etc.} (proxies for sensitive attributes) while debiasing downstream task models. Because of incomplete disentanglement in the latent code, sensitive information encoded in proxy attributes may not exist only in those removed dimensions but also in the remaining dimensions. This results in fairness degradation of downstream task models.
\end{itemize}
% \vspace{-5pt}
In this work, we aim to address the aforementioned two challenges by exploring methods that do not rely on complete disentanglement and can better cover sensitive information. To this end, we propose a novel fairness framework named DVGE (\textbf{D}ebiasing \textbf{v}ia \textbf{G}radient-based \textbf{E}xplanation), as depicted in Figure~\ref{system-training}. Specifically, to address the first challenge, DVGE does not remove latent code dimensions, which causes problems in locating sensitive information and debiasing downstream task models. To locate sensitive information and simultaneously address the second challenge, we exploit gradient-based explanations to highlight the importance of each latent code dimension when a model predicts sensitive attributes using the latent code. To debias downstream task models, we propose to perturb the latent code with the model focuses derived from gradient-based explanations. Overall, our main idea is to exploit gradient-based explanation to 1) obtain the model focus for predicting sensitive attributes, which we refer to as sensitive focus, and 2) obtain the model focus for predicting downstream task attributes, which we refer to as downstream task focus, and 3) use the two focuses to guide the training of downstream task models. Specifically, we propose \emph{bidirectional perturbation} which uses the downstream task focus to positively perturb the latent code so that models pay more attention to downstream task information, and uses the sensitive focus to reversely perturb the latent code so that models pay less attention to sensitive information.

Compared with methods based on adversarial training, DVGE is more flexible, because it separates encoding and debiasing, so that the encoder does not need retraining when sensitive attributes or downstream tasks are changed. DVGE is also less tricky to train, since it debiases via perturbation instead of adversary. Compared with methods based on disentangled representation learning, DVGE better covers sensitive information with XAI explanations and reduces useful information loss without removing latent code dimensions.

As for evaluation, we conduct extensive experiments to compare our framework with previous state-of-the-art approaches by considering disentangled and non-disentangled VAE-based representation learning methods, on both real-world unstructured dataset (CelebA~\cite{liu15faceattributes}) and structured dataset (South German Credit~\cite{groemping19south}). We measure the extent of fairness with two standard metrics, demographic parity (DP)~\cite{dwork12fairness,verma18fairness} and equal opportunity (EO)~\cite{hardt16equality}, against model accuracy. The results show that DVGE achieves better fairness-accuracy trade-off than the state-of-the-art approaches.

Our contributions are summarized as follows.
\vspace{-5pt}
\begin{itemize}
\item
We propose a novel fairness framework DVGE, to address the problem of the loss of useful downstream task information and the problem of overlooking sensitive information from proxy attributes, when debiasing models with incompletely disentangled latent code.

% Our framework copes with the sensitive information from both sensitive attributes and proxy attributes while boosting the accuracy of downstream task models without sacrificing flexibility.

\item
We introduce to exploit gradient-based explanation to obtain model focuses related to sensitive information and downstream task information, and propose \emph{bidirectional perturbation} to guide the model training for fairness purpose with the focuses.
\item
By extensive experiments, we show that our framework leads to better fairness-accuracy trade-off on both unstructured and structured real-world datasets compared to previous state-of-the-art approaches.
\end{itemize}

% \vspace{-15pt}
\section{Related Work}
\label{related}
% \vspace{-7pt}

In this section, we review the works related to our paper, namely, debiasing methods in machine learning, variational autoencoders, and gradient-based explanations.

\subsection{Debiasing Methods in Machine Learning}
% \vspace{-7pt}

The methods for debiasing can be categorized as pre-processing methods, in-processing methods, and post-processing methods. Pre-processing methods aim for generating unbiased data for training by transforming the input data. Many recent pre-processing methods focused on learning discrimination-free encodings or embeddings for various tasks~\cite{lahoti19ifair,creager19flexibly,park21learning}. As for in-processing methods, they try to remove discrimination from models during training via objective functions, fairness constraints, or through adversarial training~\cite{ganin16domain}. Furthermore, post-processing methods are proposed to audit predictions and may reassign labels with regard to fairness measurements after the training process~\cite{cui21towards}. Our proposed framework falls into the category of pre-processing methods.

% \vspace{-7pt}
\subsection{Variational Autoencoders (VAEs)}
% \vspace{-7pt}

VAEs are exploited to generate new unseen data that complies with the original distribution for generation tasks~\cite{oussidi18deep}. The main idea of VAEs is to learn a Gaussian distribution from training data and force the decoded data to have a similar distribution. Following the vanilla VAE~\cite{kingma13auto}, many variations of VAEs are proposed for different purposes, such as disentanglement~\cite{higgins16beta,kim18disentangling}, recommendation~\cite{li17collaborative}, fairness~\cite{creager19flexibly,park21learning}, etc. In this paper, we demonstrate that our fairness framework achieves better fairness-accuracy trade-off by considering both non-disentangled VAE (VanillaVAE~\cite{kingma13auto}) and disentangled VAE (FactorVAE~\cite{kim18disentangling}).

% \vspace{-7pt}
\subsection{Gradient-based Explanations}
% \vspace{-7pt}

\begin{figure*}[!t]
    \centering
    \includegraphics[width=0.9\textwidth]{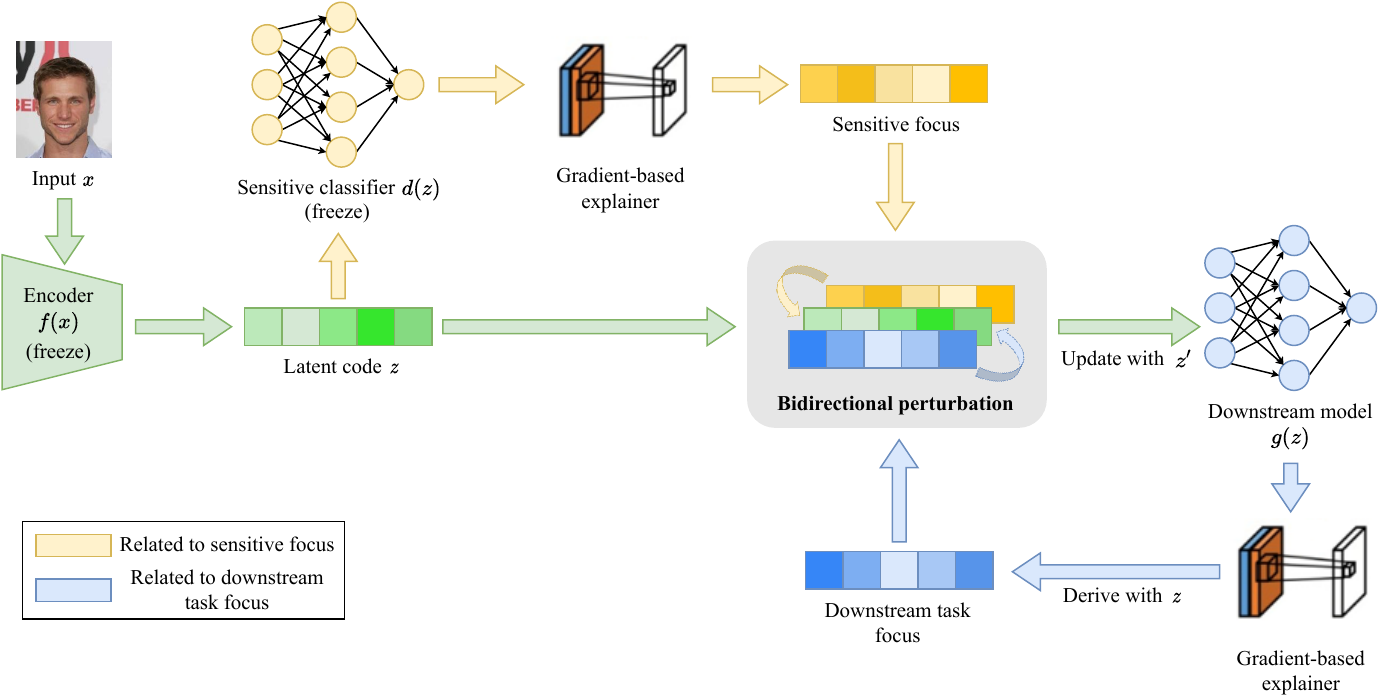}
    % \vspace{-10pt}
    \caption{The overview of DVGE training procedure. First, the latent code $z$ is generated with a trained VAE. Then, by feeding $z$ to a trained sensitive classifier and the downstream task model being trained, the sensitive focus and the downstream task focus are derived from the gradient-based explanations on them. After the bidirectional perturbation perturbs $z$ with the sensitive focus and the downstream task focus, the perturbed latent code $z^{\prime}$ is used to update the downstream task model.}
    \label{system-training}
    % \vspace{-10pt}
\end{figure*}

Gradient-based explanation methods are one of the primary approaches to explaining machine learning models, along with prototype-based methods, perturbation-based methods, etc. They produce local explanations for individual data points. The generated explanation (also known as saliency map or sensitivity map) by exploiting input-gradient highlights which parts in an input data point the model focuses on for making the prediction. Such an attempt is first made by Simonyan et al.~\cite{simonyan13deep}. Following their work, a number of variations are proposed, such as Grad-CAM~\cite{selvaraju17grad}, SmoothGrad~\cite{smilkov17smoothgrad}, FullGrad~\cite{srinivas19full}. In this work, we leverage gradient-based explanations to perturb the latent code for boosting the fairness and accuracy of downstream task models.

% \vspace{-7pt}
\section{Background}
\label{background}
% \vspace{-7pt}
Here, we briefly introduce the background of two group fairness notions that we consider in this paper.

% and previous debiasing approaches using disentangled representation learning which we treat as the baselines in this study.

% \subsection{Group Fairness Notions}
% \label{subsec:group_fairness}

In this paper, we consider two commonly used group fairness notions, demographic parity (DP)~\cite{dwork12fairness,verma18fairness} and equal opportunity (EO)~\cite{hardt16equality}. Let us first consider a simple example, in which we train a model $\hat{y}=g(x)$ to predict the label $y\in\{0,1\}$, where $\hat{y}$ is the prediction, $x$ denotes the input data, $s\in\{s_{1},s_{2}\}$ denotes sensitive attributes in the input.

\textbf{Demographic Parity.} The definition of DP is that the model prediction is independent of sensitive attributes. In other words, the probability that a member of any subgroup ($s_{1}$ or $s_{2}$) receives the same prediction, 0 or 1 in our example, is completely the same. Based on the definition, the distance to demographic parity $\Delta_{DP}$ is used to measure how fair a model is as
\begin{equation}
\Delta_{DP}=|P(\hat{y}=1|s=s_{1})-P(\hat{y}=1|s=s_{2})|.
\end{equation}
When $\Delta_{DP}=0$, the demographic disparity is satisfied.

\textbf{Equal Opportunity.} Equal opportunity indicates that the true positive rate(TPR) of a model remains the same with respect to each subgroup. This is mathematically equivalent to that each subgroup has the same false negative rate (FNR). We can also use the distance to EO $\Delta_{EO}$ to measure the extent of fairness of a model as
\begin{equation}
\Delta_{EO}=|P(\hat{y}=1|s=s_{1}, y=1)-P(\hat{y}=1|s=s_{2},y=1)|
\end{equation}
or
\begin{equation}
\Delta_{EO}=|P(\hat{y}=0|s=s_{1}, y=1)-P(\hat{y}=0|s=s_{2},y=1)|.
\end{equation}
This definition underlines the idea that the qualified members of each subgroup should have the same probability to receive positive or negative predictions.

\section{The Proposed Fairness Framework}
\label{framework}
% \vspace{-7pt}

% As we discussed before, since previous approaches using disentangled representation learning suffer from incomplete disentanglement and debiases models via removing the latent code dimensions related to sensitive attributes, they have two major resulted limitations: 1) the loss of the useful information for downstream tasks causes accuracy degradation of downstream task models, and 2) the overlook of sensitive information carried by proxy attributes leads to fairness degradation of downstream models.

In this paper, we design a new fairness framework DVGE, as illustrated in Figure~\ref{system-training}, by considering sensitive information from both sensitive attributes and proxy attributes. The framework does not depend on complete disentanglement. Instead, it leverages gradient-based explanation to obtain model focuses for predicting sensitive attributes and downstream task labels, and uses the proposed bidirectional perturbation to perturb the latent code for guiding the training of downstream task models.

% In this section, we first briefly introduce the process of learning the latent code. Second, we elaborate obtaining sensitive focus and downstream task focus with gradient-based explanation. Third, we introduce the proposed bidirectional perturbation.
% \vspace{-15pt}
\subsection{Latent Code}
% \vspace{-7pt}

\begin{figure*}[!t]
    \centering
    \includegraphics[width=0.9\textwidth]{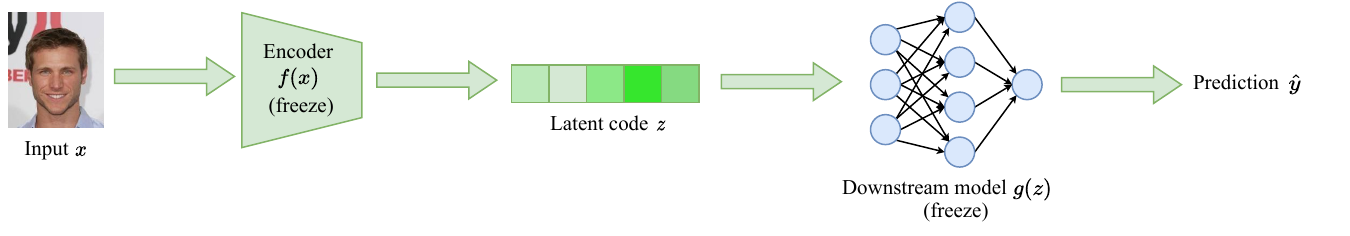}
    % \vspace{-10pt}
    \caption{The overview of DVGE inference procedure. After training, DVGE does not perform bidirectional perturbation on the latent code during inference.}
    \label{system-prediction}
    % \vspace{-10pt}
\end{figure*}

As our work follows the idea of using representation learning to debias machine learning models with the flexibility to cope with different sensitive attributes and downstream tasks, we first train a VAE $f(x)$ to encode the input data $x$ into latent code $z$ by maximizing the Evidence Lower Bound (ELBO)~\cite{kingma13auto}. The VAE used in DVGE is fixed after training. Our framework works with both disentangled and non-disentangled VAEs, which we show in the experiments.
% by testing our framework with VanillaVAE~\cite{kingma13auto} (non-disentangled) and FactorVAE~\cite{kim18disentangling} (disentangled).

% \vspace{-12pt}
\subsection{Sensitive Focus}
\label{sensitive_focus}
% \vspace{-7pt}
Sensitive focus is the model focus for predicting sensitive attributes $s$ with latent code $z$. We derive it using a gradient-based explanation, since this explanation is input-specific and assigns an importance score to each latent code dimension based on gradients, which can be easily used for perturbation. Given a trained sensitive classifier $d(z)$ which takes the latent code $z$ as input to predict sensitive attributes $s$, the gradient-based explanation $e_{sens}$ for its prediction is calculated as
\begin{equation}
    e_{sens}=\psi(\nabla_{z}d(z)\odot z),
\end{equation}
where $\psi(\cdot)$ is a post-processing operation for a gradient-based explanation, e.g., scale and taking the absolute value, $\nabla_{z}d(z)$ is the model gradient with regard to $z$, and $\odot$ denotes element-wise multiplication. As $\psi(\cdot)$ and $\odot$ are only for the visualization purpose of the explanation, the essence of the explanation is $\nabla_{z}d(z)$, we define the sensitive focus as
\begin{equation}
    F_{sens}=\nabla_{z}d(z).
\end{equation}
Please note that $\nabla_{z}d(z)$ is computed via backpropagation with the predicted sensitive attributes $\hat{s}=d(z)$. Thus, computing sensitive focus does not require access to real sensitive attributes.

\textbf{Sensitive Information Coverage.} Since the sensitive classifier $d(z)$ is trained to make use of every dimension of the latent code $z$ to make predictions about sensitive attributes $s$, any sensitive information or shortcut information linking to $s$ is exploited by it for the prediction. In addition, the gradient-based explanation can highlight all this information in \textbf{every dimension} of $z$, so the defined sensitive focus in our framework covers the sensitive information from both sensitive attributes and proxy attributes in the latent code.

% As a result, in the SCM shown in Figure~\ref{scm_2}, our framework leverages the sensitive focus to break both $s\rightarrow z$ and $p\rightarrow z$.

\textbf{Flexibility w.r.t. Changing Sensitive Attributes.} As the sensitive focus is obtained via gradient-based explanation on sensitive classifier $d(z)$, when different sensitive attributes are required, we only need to change to a new $d(z)$ for predicting the new version of $s$, while reusing the latent code $z$.

% \vspace{-12pt}
\subsection{Downstream Task Focus}
% \vspace{-7pt}

The downstream task focus is the model focus for predicting downstream task label $y$ with the latent code $z$. We obtain this focus directly from the gradient-based explanation of the downstream task model during its training process. Similar to Section~\ref{sensitive_focus}, given a downstream task model $g(z)$ and the latent $z$, the gradient-based explanation of the model is calculated as
\begin{equation} e_{task}=\psi(\nabla_{z}g(z)\odot z),
\end{equation}
and we define the downstream task focus as
\begin{equation}
F_{task}=\nabla_{z}g(z).
\end{equation}
The downstream task focus is for boosting the accuracy performance of the downstream task model while debiasing.
% \vspace{-7pt}

\begin{figure*}[!t]
    \centering
    \subfigure[DVGE-D, $\Delta_{DP}$]{
    \includegraphics[width=0.235\textwidth]{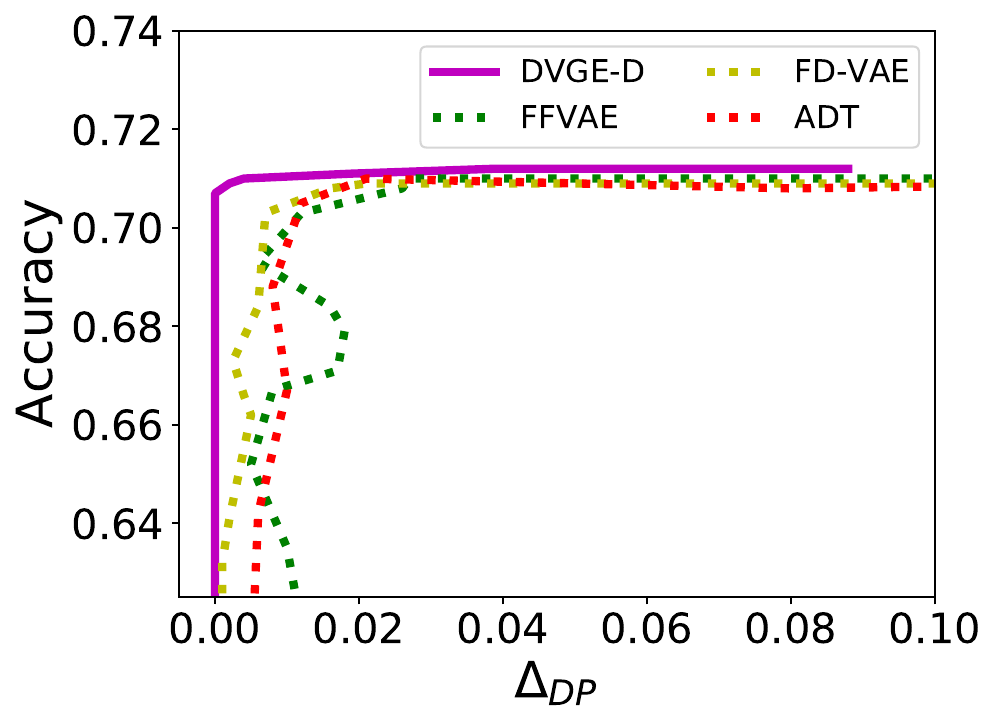}
    }
    \subfigure[DVGE-D, $\Delta_{EO}$]{
    \includegraphics[width=0.235\textwidth]{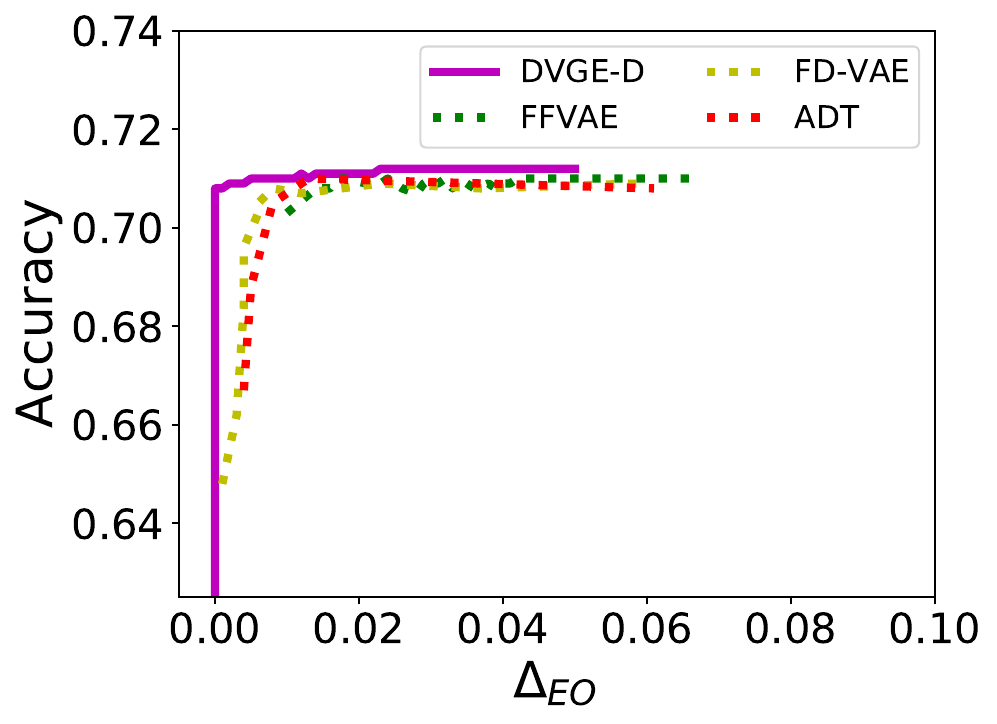}
    }
    \subfigure[DVGE-N, $\Delta_{DP}$]{
    \includegraphics[width=0.235\textwidth]{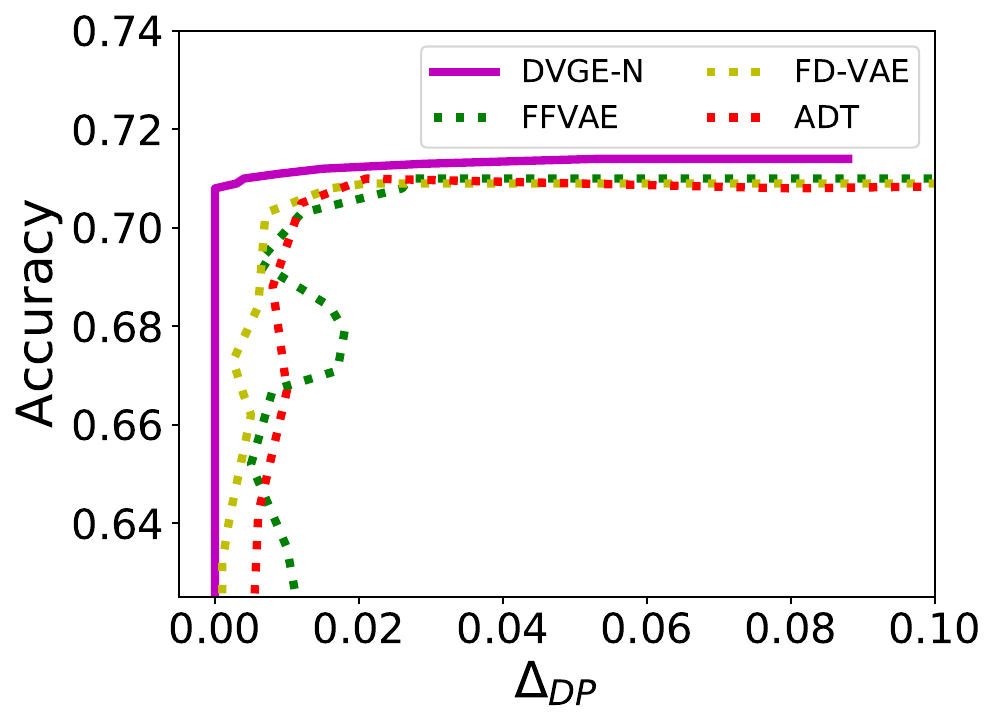}
    }
    \subfigure[DVGE-N, $\Delta_{EO}$]{
    \includegraphics[width=0.235\textwidth]{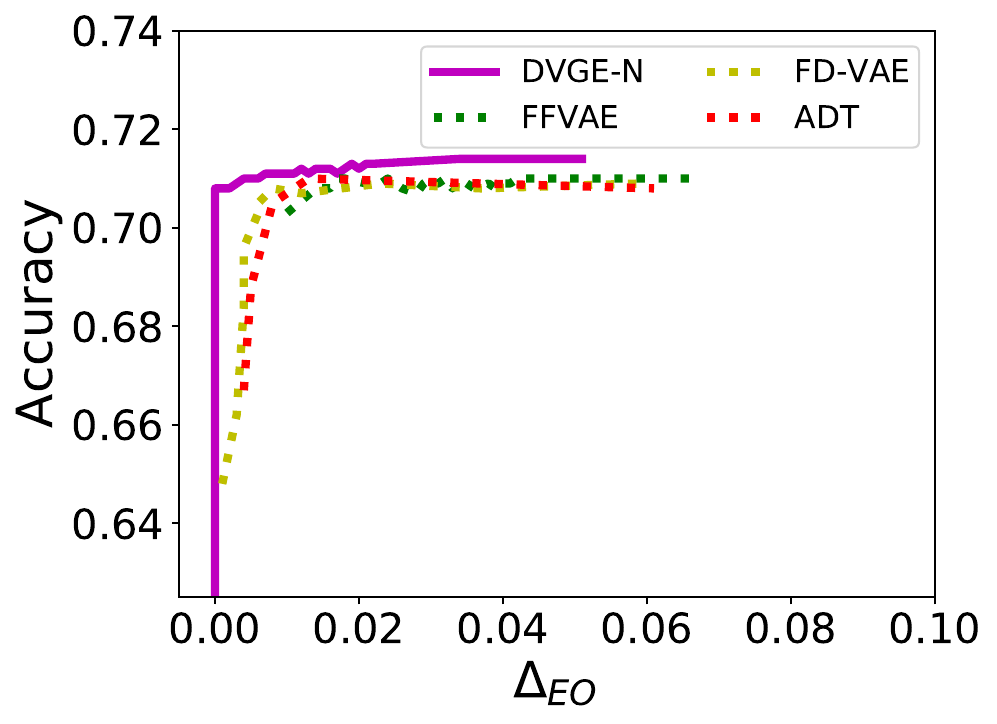}
    }
    % \vspace{-9pt}
    \caption{Fairness-accuracy trade-off comparison results for Experiment 1: CelebA dataset, sensitive attribute = ``Male", task label = ``Oval\_Face".}
    \label{experiment_1}
    % \vspace{-10pt}
\end{figure*}

\subsection{Bidirectional Perturbation}
% \vspace{-7pt}
We perform bidirectional perturbation by perturbing the latent code $z$ with the sensitive focus $F_{sens}$ and the downstream task focus $F_{task}$ to guide the training of the downstream task model for the purpose of fairness and prevention of downstream task accuracy degradation. The perturbed latent code $z^{\prime}$ is calculated as
\begin{equation}
\begin{split}
z^{\prime} &= z + Clip_{\epsilon}\{\eta_{1}*F_{sens}-\eta_{2}*F_{task}\} \\
&=z + Clip_{\epsilon}\{\eta_{1}*\nabla_{z}d(z)-\eta_{2}*\nabla_{z}g(z)\},
\end{split}
\label{eq:bp}
\end{equation}
where $\eta_{1}$ and $\eta_{2}$ are the hyperparameters for controlling the intensity of debiasing and accuracy boosting, respectively, and
\begin{equation}
Clip_{\epsilon}\{v\}=\begin{cases}
    \epsilon, \text{ if }v>\epsilon,\\
    max(v, -\epsilon), \text{ otherwise},
\end{cases}
\end{equation}
where $\epsilon$ is non-negative and denotes the threshold for the distortion caused by bidirectional perturbation. $Clip_{\epsilon}\{\cdot\}$ is designed to prevent bidirectional perturbation introducing too much information distortion on latent code dimensions.

\textbf{Rationale behind Bidirectional Perturbation}. Since backpropagation gradients indicate the directions to optimize the objective function, in the equation~\ref{eq:bp}, $+\nabla_{z}d(z)$ (sensitive focus) is for updating the latent code $z$ in the reverse direction of optimizing sensitive information prediction, while $-\nabla_{z}g(z)$ (downstream task focus) is for updating $z$ towards the direction of optimizing downstream task model, so that the downstream task model is guided to pay less attention to sensitive information and more attention to downstream task information. DVGE uses the perturbed latent code $z^{\prime}$ to update the downstream task model.

\textbf{No Reliance on Complete Disentanglement.} Even if the factors of variation for sensitive information are not fully disentangled (mixed with other factors of variation in the latent code dimensions), our framework still can debias the downstream task model, since the sensitive focus covers the sensitive information in \emph{every dimension} of the latent code $z$, and our framework leverages the sensitive focus to perturb \emph{every dimension} of $z$.

\textbf{Inference.} After finishing training the downstream task model with our framework, we do not perform the bidirectional perturbation on the latent code during inference as shown in Figure~\ref{system-prediction}. The reason is that the model after training has learned to pay more attention to downstream task information and less attention to sensitive information in the latent code.
% \vspace{-7pt}

\section{Experiments}
\label{experiments}
% \vspace{-7pt}

We conduct extensive experiments to evaluate our framework while comparing it with previous state-of-the-art approaches. To show the flexibility of our framework, we consider different sensitive attributes individually and jointly on structured dataset and unstructured dataset. To demonstrate that our framework does not rely on complete disentanglement, we consider both non-disentangled and disentangled VAEs. Furthermore, we use an ablation study to demonstrate that our framework has better coverage on sensitive information.

% via sensitive focus than the previous approaches which remove latent code dimensions correlated with sensitive attributes. In this section, we first elaborate on the experiment setups, and then present the experiment results.

% \vspace{-12pt}
\subsection{Experiment Setups}
% \vspace{-3pt}

\subsubsection{DVGE-D and DVGE-N}

To demonstrate that our framework does not rely on complete disentanglement to debias downstream task models, we implement DVGE with one disentangled VAE (FactorVAE~\cite{kim18disentangling}) and one non-disentangled VAE (VanillaVAE~\cite{kingma13auto}), respectively. And we denote them as \textbf{DVGE-D} and \textbf{DVGE-N}. For more implementation details, please refer to Appendix~\ref{implementation}.

% There are basically three steps to implement DVGE. First, we train VAEs to produce the latent code. Then, we train a sensitive classifier with the latent code. Finally, we train the downstream task model with the latent code according to our framework. For more implementation details, please refer to Appendix~\ref{implementation}.

% \vspace{-15pt}
\subsubsection{Baselines}

We consider three state-of-the-art debiasing approaches as baselines in the experiments.
% \vspace{-7pt}
\begin{itemize}
    \item Adversarial Training (ADT)~\cite{ganin16domain}: The model based on ADT consists of three parts, i.e., feature encoder, sensitive branch, and downstream task branch. ADT debiases the model by updating the feature encoder with the reverse loss of the sensitive branch.
    \item FFVAE~\cite{creager19flexibly}: Based on previous disentangled representation learning methods, FFVAE tries to explicitly separate sensitive dimensions from non-sensitive dimensions in the latent code by learning the sensitive latent part with supervised learning.
    \item FD-VAE~\cite{park21learning}: FD-VAE separates the latent code into three portions, i.e., sensitive dimensions, downstream-task-related dimensions, and mutual-information dimensions. FD-VAE trains the downstream task model using the latent code without sensitive dimensions while trying to exclude sensitive information from mutual-information dimensions with adversarial training.
\end{itemize}
% \vspace{-7pt}
Before training the encoder, ADT and FD-VAE require to specify the sensitive attributes and the downstream task attribute, while FFVAE requires to specify the sensitive attributes. In contrast, our framework does not require to specify either of them and has the highest flexibility. Since the debiasing process in our framework is not based on adversarial training, DVGE is more stable and easier to train than the baselines.

% \vspace{-12pt}
\subsubsection{Datasets}

In the experiments, we use two commonly used datasets. One is an unstructured dataset, which is CelebA\footnote{\url{https://mmlab.ie.cuhk.edu.hk/projects/CelebA.html}}~\cite{liu15faceattributes}, while the other is a structured dataset, which is South German Credit\footnote{\url{https://archive.ics.uci.edu/ml/datasets/South+German+Credit}}~\cite{groemping19south}. CelebA has 202,599 facial images, each of which is associated with 40 attributes, such as ``Attractive", ``Male", ``Young". And all attributes are in binary form. As for the structured dataset, South German Credit has 1,000 entries with 21 attributes. The first 20 attributes are the information about the loan applicants (gender, age, income, etc.), and the last one is the loan application result. Since some attributes in South German Credit are in category form, we convert them into numerical form for convenience.

% \vspace{-9pt}
\subsubsection{Metrics}

In the experiments, we compare our framework with the baselines on the fairness-accuracy trade-off. Specifically, we consider two common fairness metrics, the distance to demographic parity $\Delta_{DP}$ and the distance to equal opportunity $\Delta_{EO}$ (refer to Section~\ref{background}). We calculate the fairness metrics against the accuracy (Acc.) of the downstream task model, and plot the Pareto fronts of them to show the fairness-accuracy trade-off. \textbf{Better fairness-accuracy trade-off indicates higher accuracy with lower $\Delta_{DP}$ and $\Delta_{EO}$}. In order to obtain the fairness-accuracy trade-off for our framework and the baselines, we sweep a range of the value combinations of hyperparameters in their objective functions.

\begin{figure*}[!t]
    \centering
    \subfigure[DVGE-D, $\Delta_{DP}$]{
    \includegraphics[width=0.235\textwidth]{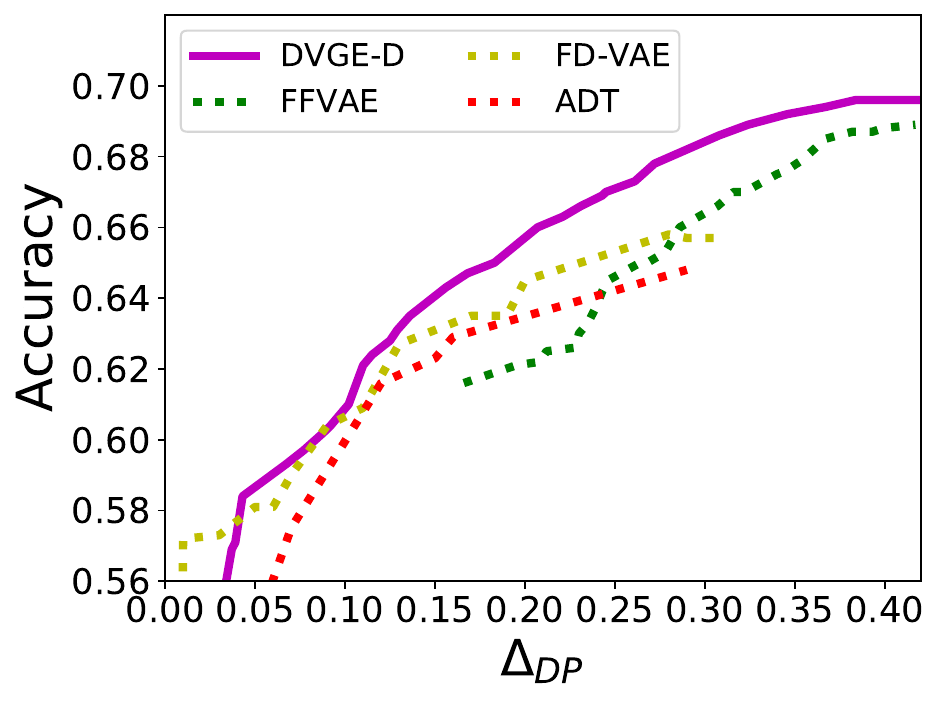}
    \label{experiment_2_1}
    }
    \subfigure[DVGE-D, $\Delta_{EO}$]{
    \includegraphics[width=0.235\textwidth]{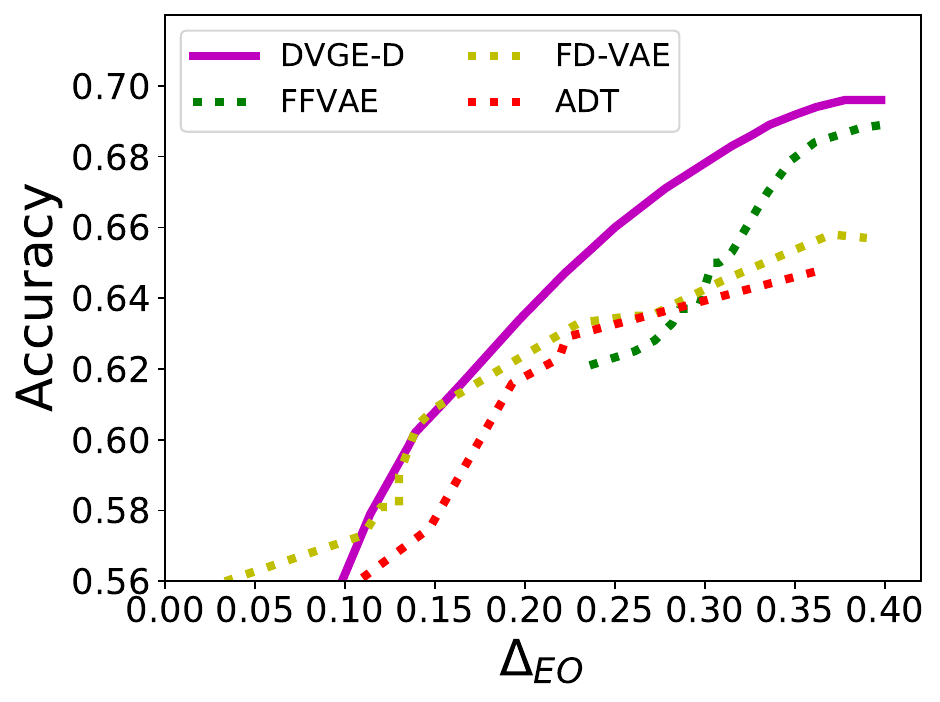}
    }
    \subfigure[DVGE-N, $\Delta_{DP}$]{
    \includegraphics[width=0.235\textwidth]{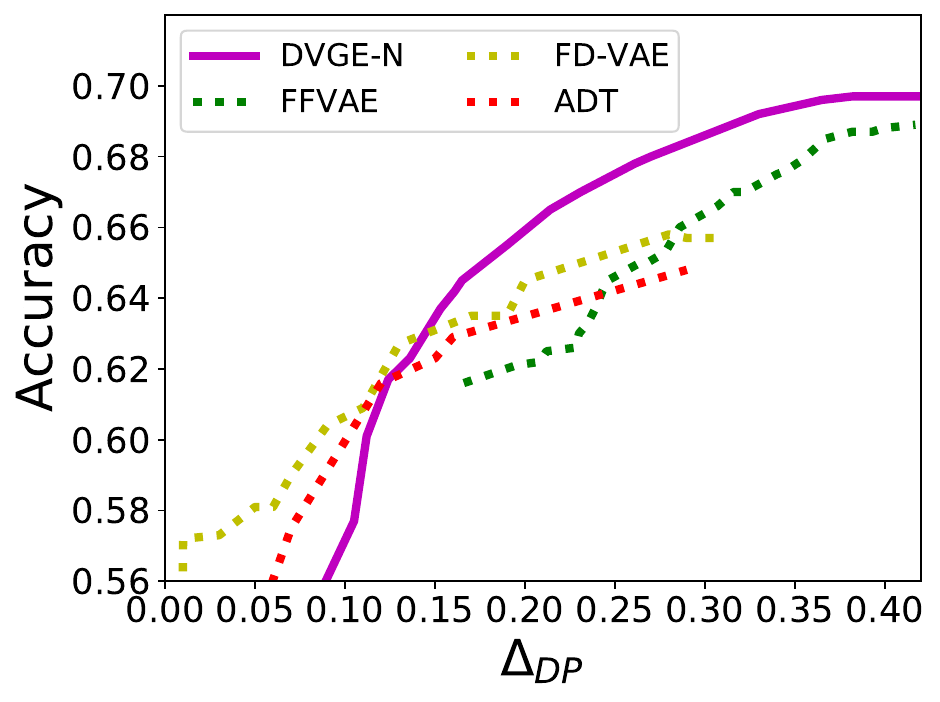}
    \label{experiment_2_3}
    }
    \subfigure[DVGE-N, $\Delta_{EO}$]{
    \includegraphics[width=0.235\textwidth]{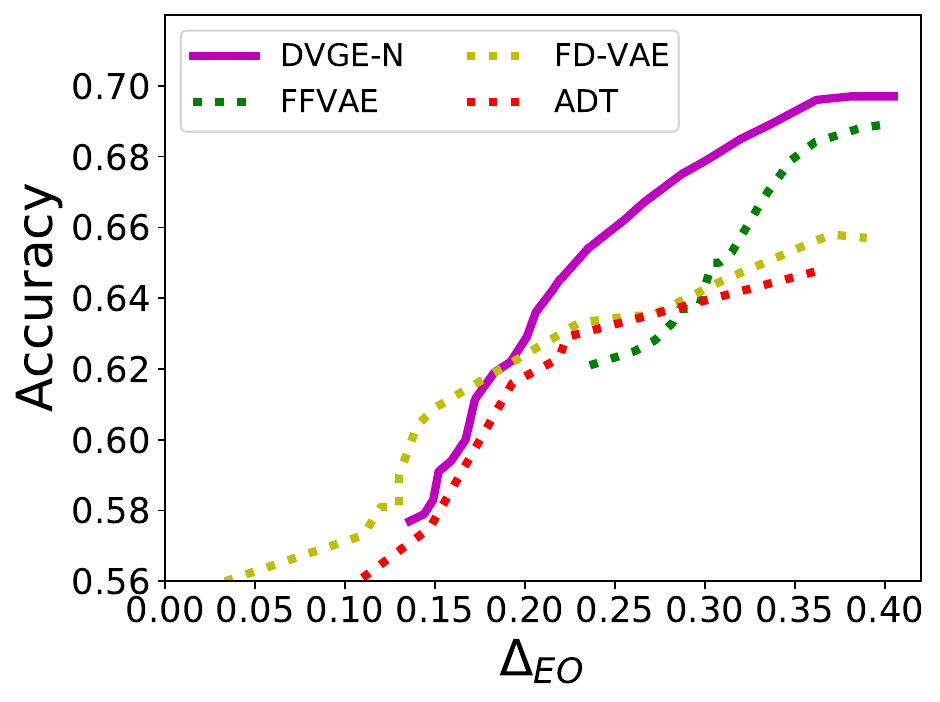}
    \label{experiment_2_4}
    }
    % \vspace{-9pt}
    \caption{Fairness-accuracy trade-off comparison results for Experiment 2: CelebA dataset, sensitive attribute = ``Male", task label = ``Attractive".}
    \label{experiment_2}
    % \vspace{-10pt}
\end{figure*}

\begin{figure*}[!t]
    \centering
    \subfigure[DVGE-D, $\Delta_{DP}$]{
    \includegraphics[width=0.235\textwidth]{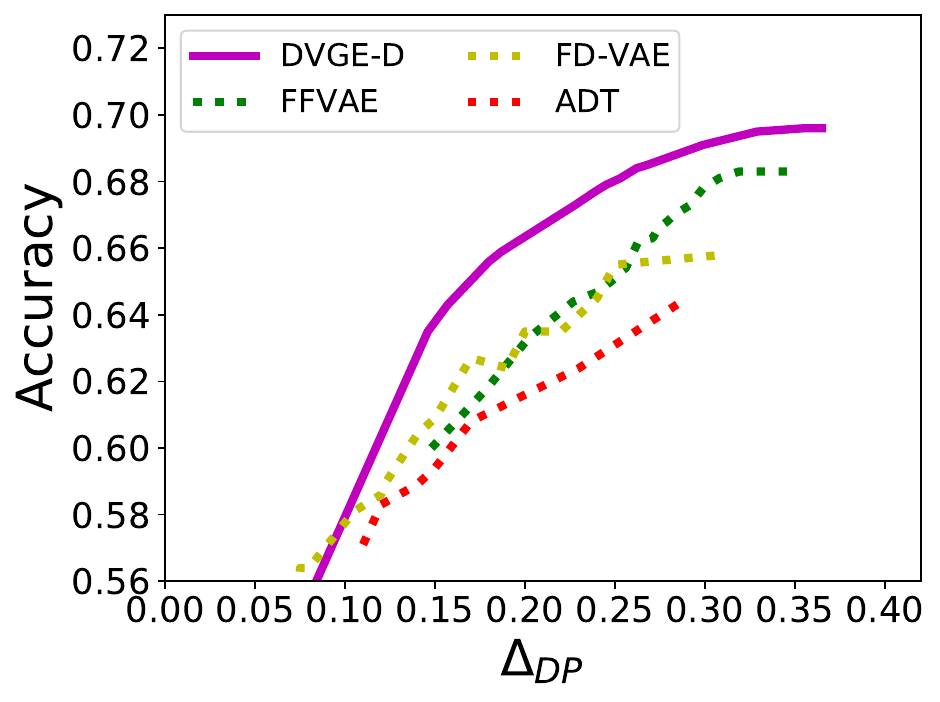}
    }
    \subfigure[DVGE-D, $\Delta_{EO}$]{
    \includegraphics[width=0.235\textwidth]{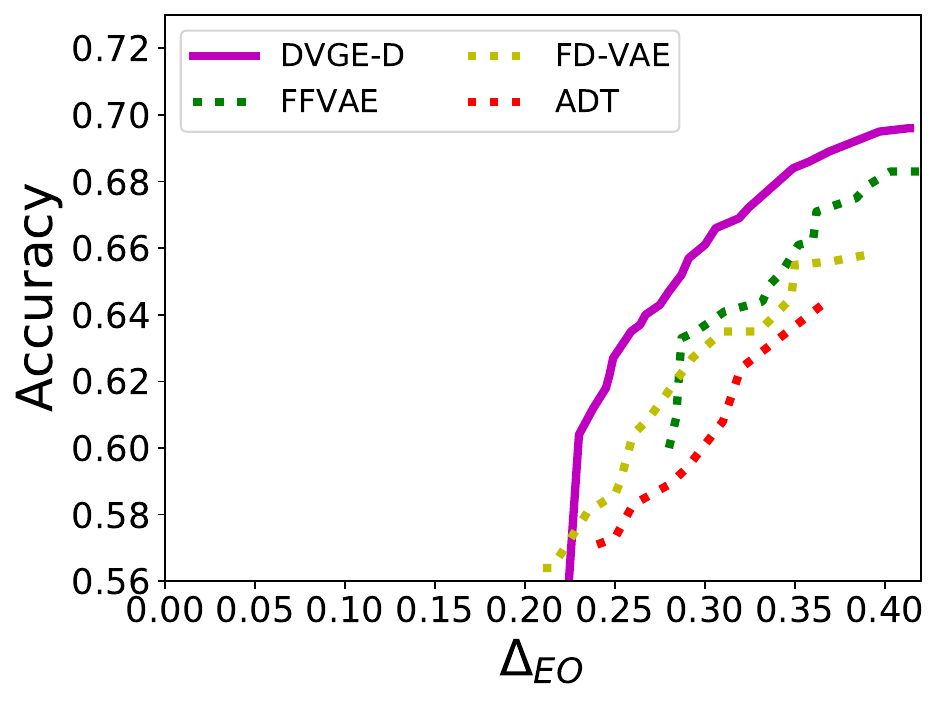}
    \label{experiment_3_2}
    }
    \subfigure[DVGE-N, $\Delta_{DP}$]{
    \includegraphics[width=0.235\textwidth]{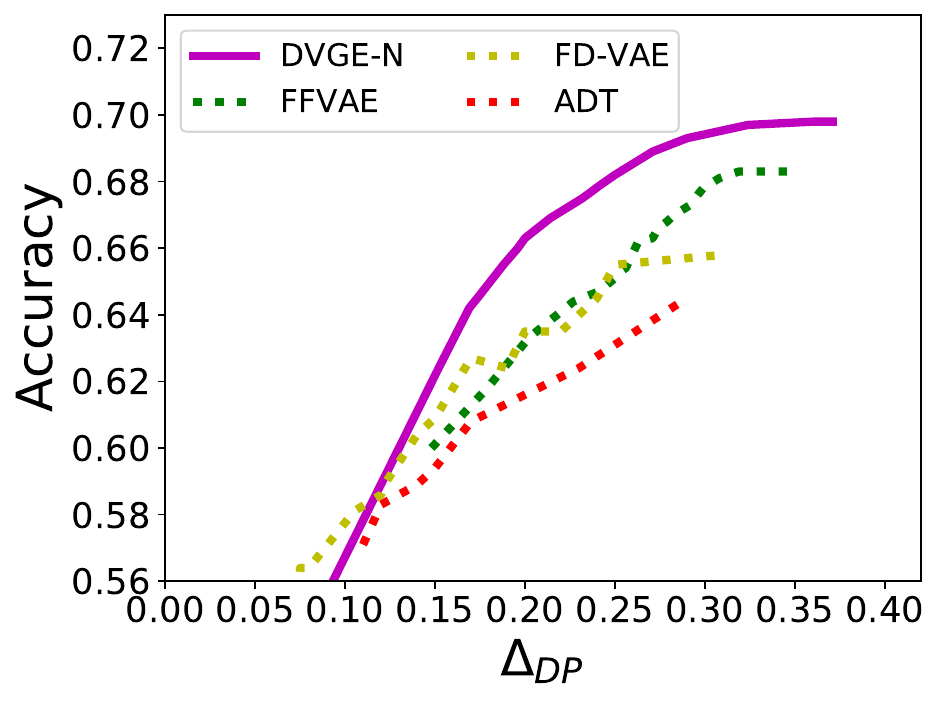}
    }
    \subfigure[DVGE-N, $\Delta_{EO}$]{
    \includegraphics[width=0.235\textwidth]{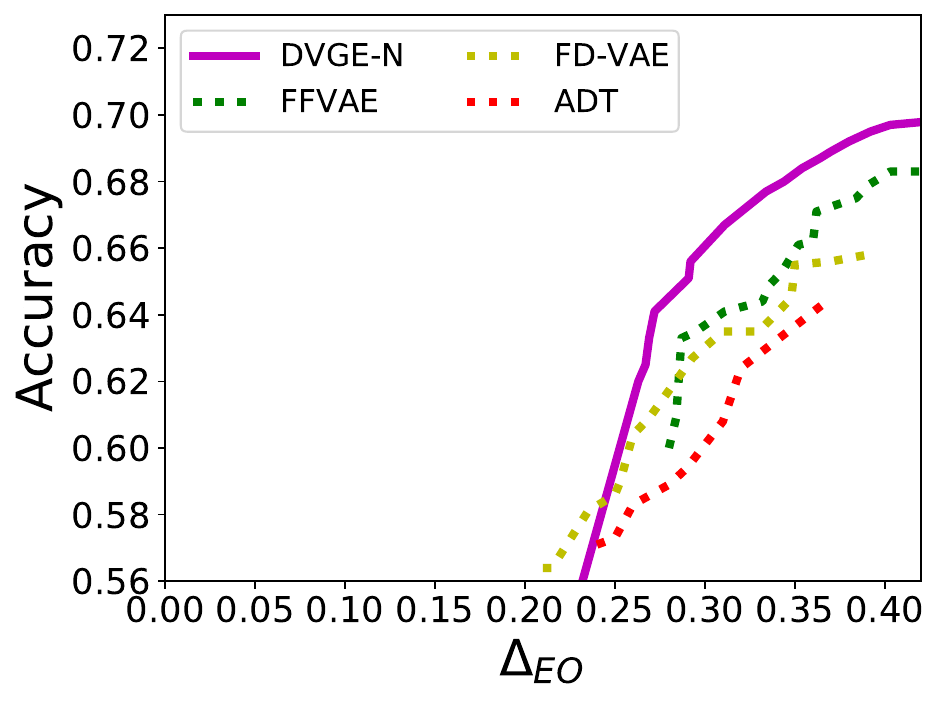}
    }
    % \vspace{-9pt}
    \caption{Fairness-accuracy trade-off comparison results for Experiment 3: CelebA dataset, sensitive attribute = ``Male" $\land $ ``Young", task label = ``Attractive".}
    \label{experiment_3}
    % \vspace{-10pt}
\end{figure*}

% \vspace{-7pt}
\subsection{Experiment Results on CelebA}
% \vspace{-3pt}

On CelebA, we select three different combinations of sensitive attributes and downstream tasks for the experiments on the unstructured dataset. Because of the flexibility, DVGE uses the same latent code encoder for the following three different experiments.

% \vspace{-15pt}
\subsubsection{Experiment 1}

We choose ``Male" as the sensitive attribute and set the downstream task to predict the label ``Oval\_Face". For this task, if we had a perfect classifier ($100\%$ accuracy), its $\Delta_{DP}$ would be 0.094, indicating almost no fairness problem in this task. When there is no fairness problem, the accuracy of the downstream task model should not vary with $\Delta_{DP}$ or $\Delta_{EO}$. This experiment is designed to verify that DVGE has no negative impacts on tasks without fairness problems.

As we can observe in Figure~\ref{experiment_1}, the accuracy of the downstream task model barely changes when $\Delta_{DP}$ or $\Delta_{EO}$ increases for our framework and the baselines. In addition, our framework can maintain the model accuracy even when $\Delta_{DP}$ and $\Delta_{EO}$ are very close to $0$. We can also observe that our framework achieves slightly better accuracy than FFVAE and FD-VAE, because they remove dimensions of the latent code and suffers from incomplete disentanglement, resulting in information loss for downstream tasks.

% \vspace{-15pt}
\subsubsection{Experiment 2}

The sensitive attribute for this experiment is also ``Male", but we change the task label to ``Attractive". $\Delta_{DP}$ for a perfect classifier in this task would be 0.398, indicating a serious fairness problem. This experiment evaluates DVGE when debiasing in the setting of single sensitive attributes.

As we can see from the experiment results in Figure~\ref{experiment_2}, our framework outperforms the baselines by a relatively large margin. For example in Figure~\ref{experiment_2_1}, DVGE-D almost always achieves higher accuracy than the baselines when at the same $\Delta_{DP}$. More importantly, our framework achieves similar fairness-accuracy trade-off with a non-disentangled VAE setting (DVGE-N in Figure~\ref{experiment_2_3} and~\ref{experiment_2_4}) as with disentangled VAE setting, which demonstrates that our framework does not rely on complete disentanglement for debiasing.

% \vspace{-15pt}
\subsubsection{Experiment 3}

In order to demonstrate the flexibility and superiority of our framework in the case of multiple sensitive attributes, we consider the conjunction of two sensitive attributes in this experiment. Specifically, the sensitive attributes are ``Male" and ``Young", denoted as ``Male" $\land $ ``Young"\footnote{$\land $ represents logical \emph{and}.}, and the task is still to predict the label ``Attractive". Here, we train a sensitive classifier to jointly distinguish the two sensitive attributes from the latent code. $\Delta_{DP}$ for a perfect classifier in this task would be 0.445, suggesting an even more serious fairness problem than those in previous tasks.

We depict the results for this experiment in Figure~\ref{experiment_3}. As we can observe, our framework overall achieves better fairness-accuracy trade-off than the baselines. For example in Figure~\ref{experiment_3_2}, when achieving the same $\Delta_{EO}$, DVGE-D always hits higher downstream task accuracy than other baselines. Even when $\Delta_{DP}$ or $\Delta_{EO}$ moves close to 0, and the gaps of the fairness-accuracy trade-off between our framework and the baselines get smaller, our framework still outperforms or is on par with the baselines.

\begin{figure*}[!t]
    \centering
    \subfigure[DVGE-D, $\Delta_{DP}$]{
    \includegraphics[width=0.235\textwidth]{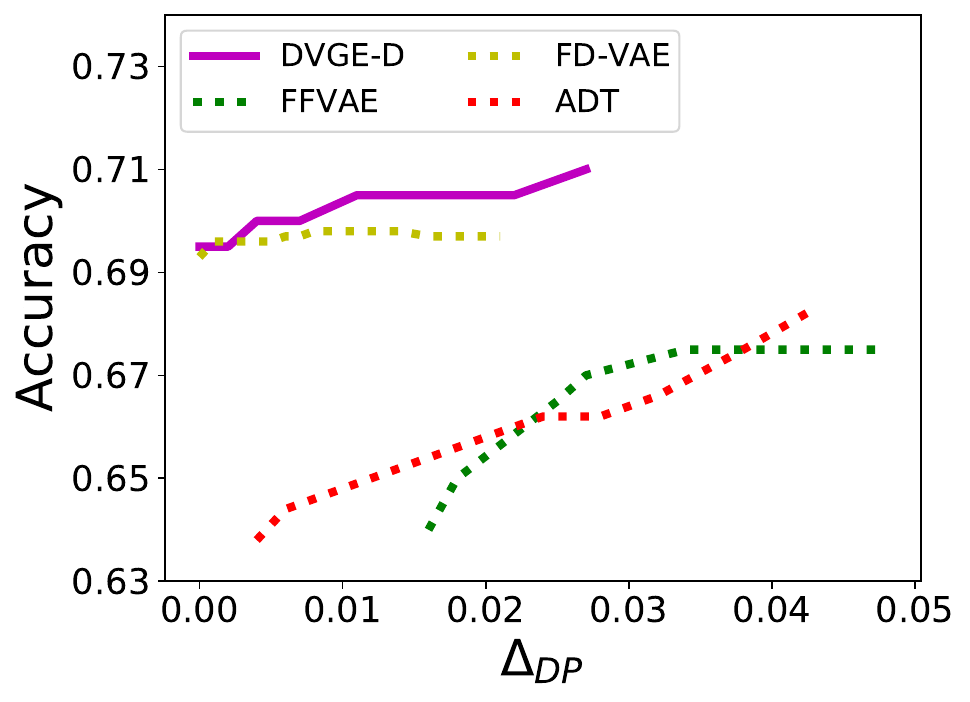}
    }
    \subfigure[DVGE-D, $\Delta_{EO}$]{
    \includegraphics[width=0.235\textwidth]{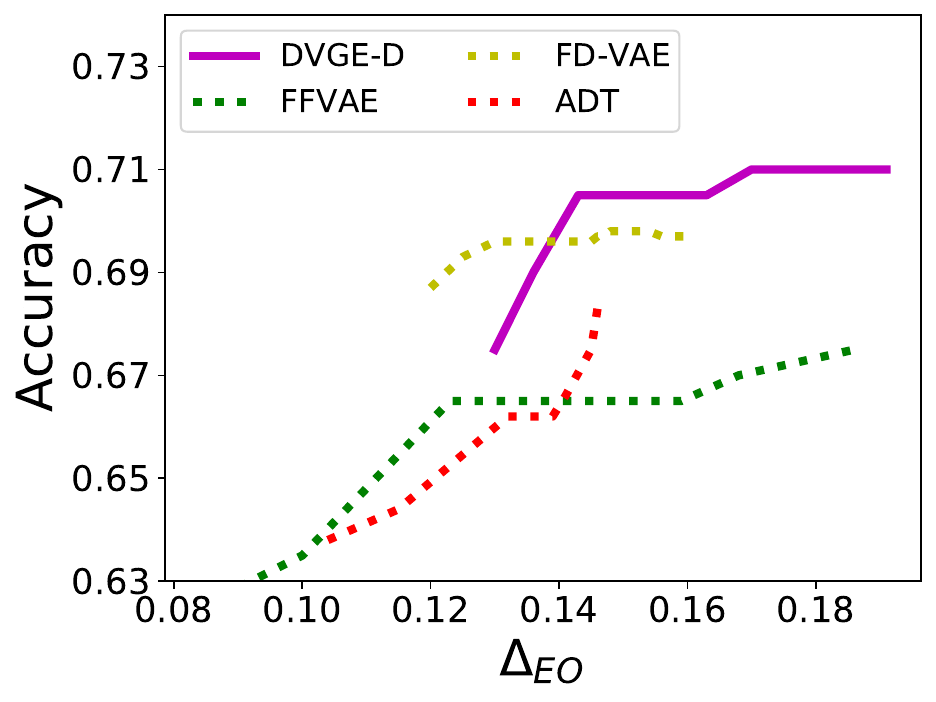}
    }
    \subfigure[DVGE-N, $\Delta_{DP}$]{
    \includegraphics[width=0.235\textwidth]{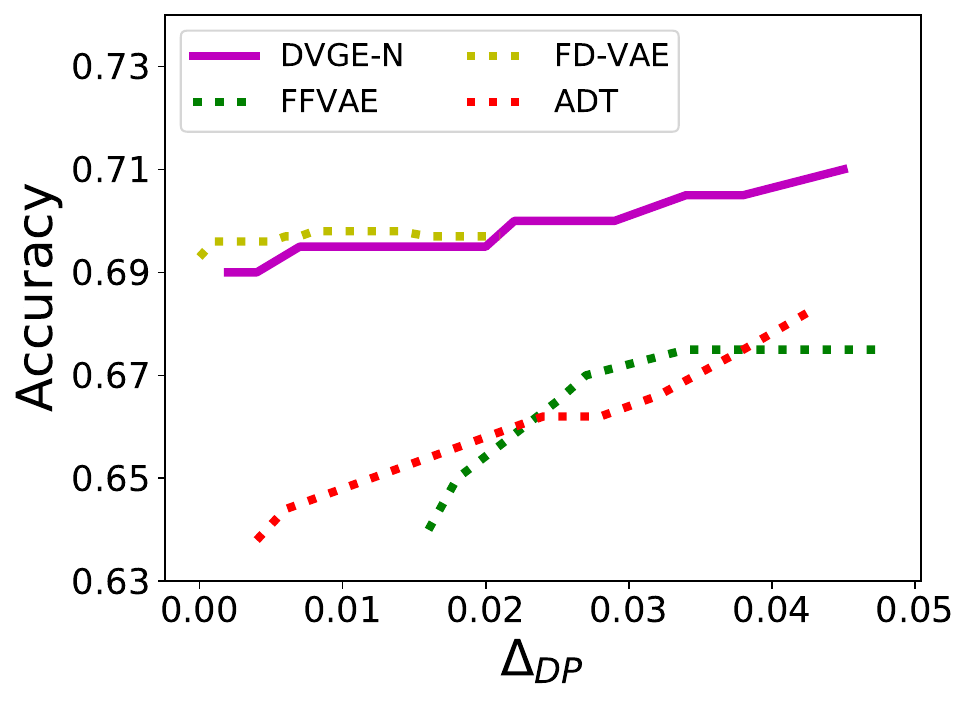}
    }
    \subfigure[DVGE-N, $\Delta_{EO}$]{
    \includegraphics[width=0.235\textwidth]{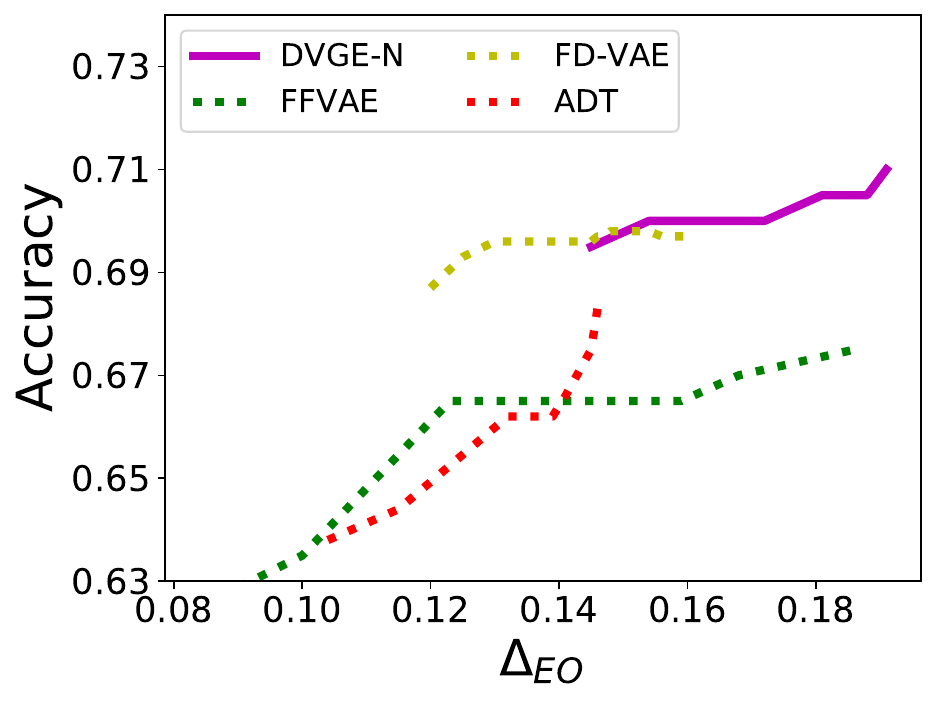}
    }
    % \vspace{-9pt}
    \caption{Fairness-accuracy trade-off comparison results for Experiment 4: South German Credit dataset, sensitive attribute = ``age", task label = ``credit\_risk".}
    \label{experiment_4}
    % \vspace{-10pt}
\end{figure*}

\begin{figure*}[!t]
    \centering
    \subfigure[DVGE-D, $\Delta_{DP}$]{
    \includegraphics[width=0.235\textwidth]{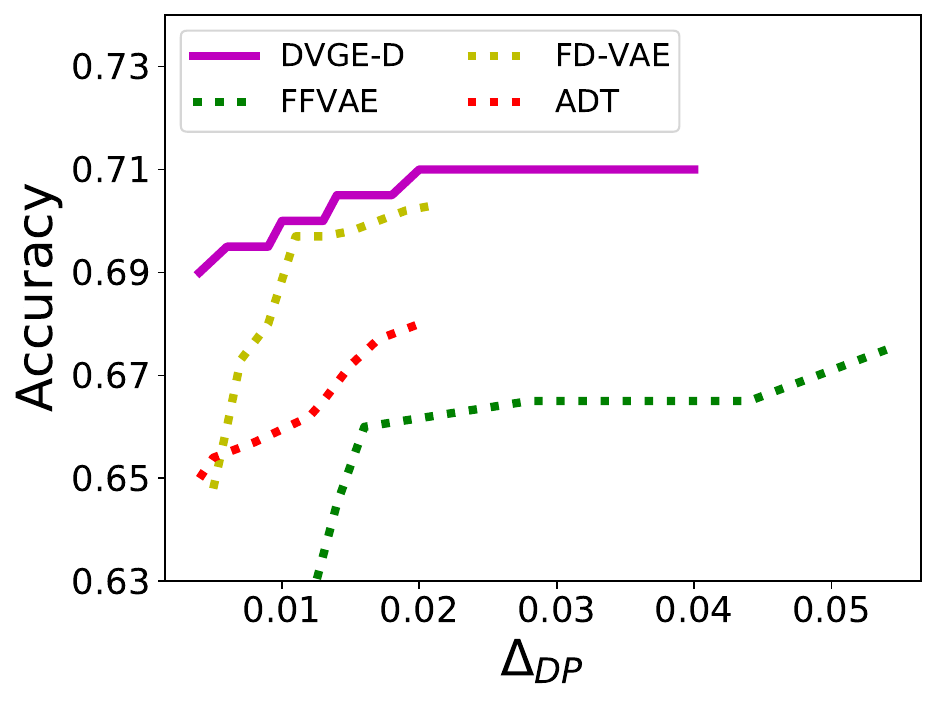}
    }
    \subfigure[DVGE-D, $\Delta_{EO}$]{
    \includegraphics[width=0.235\textwidth]{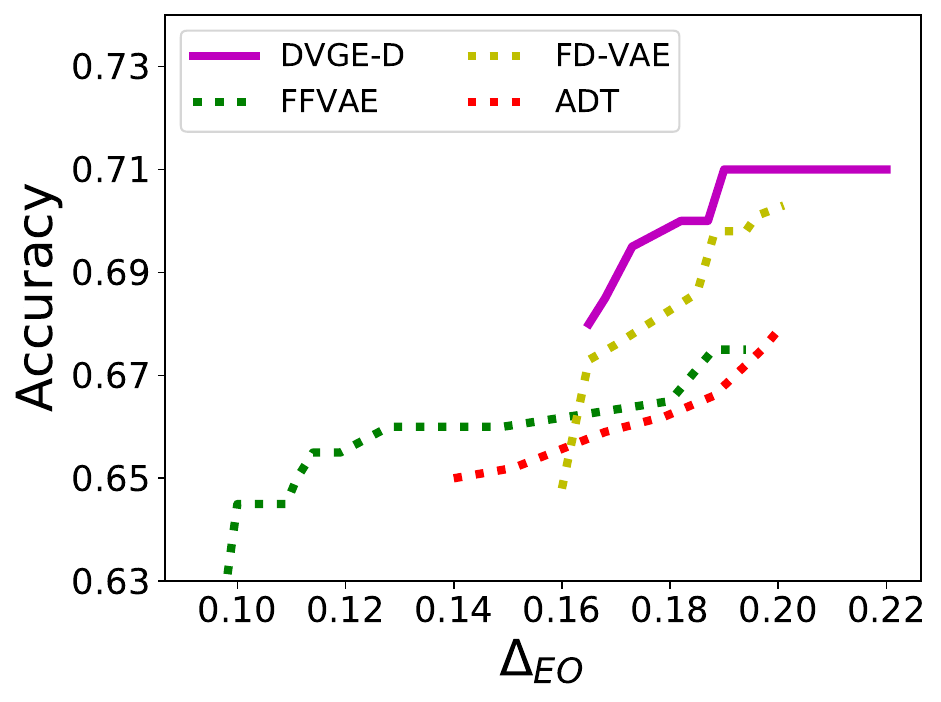}
    }
    \subfigure[DVGE-N, $\Delta_{DP}$]{
    \includegraphics[width=0.235\textwidth]{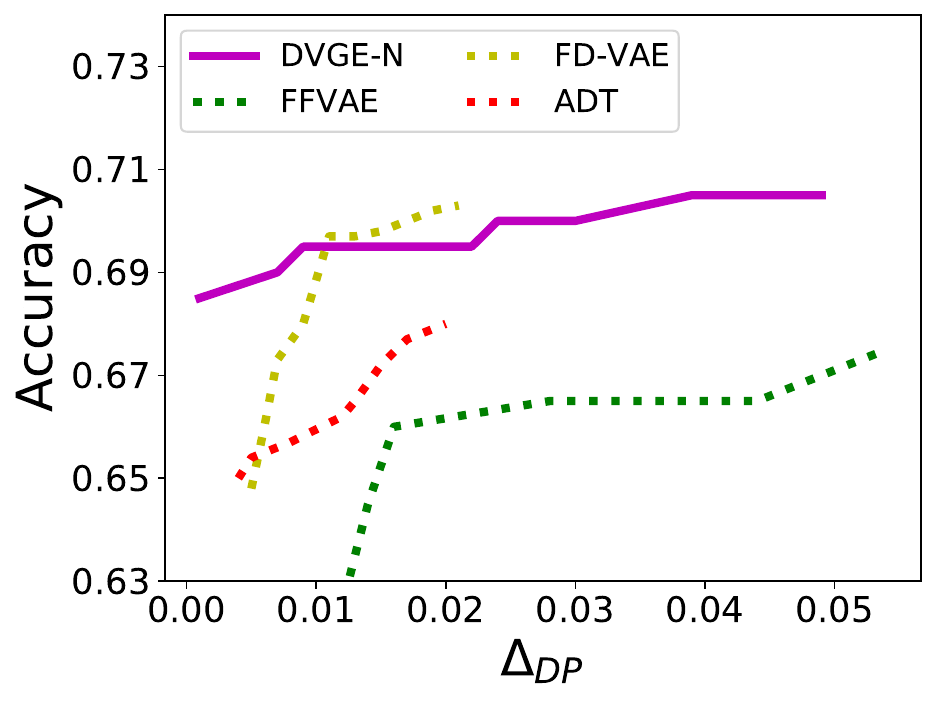}
    }
    \subfigure[DVGE-N, $\Delta_{EO}$]{
    \includegraphics[width=0.235\textwidth]{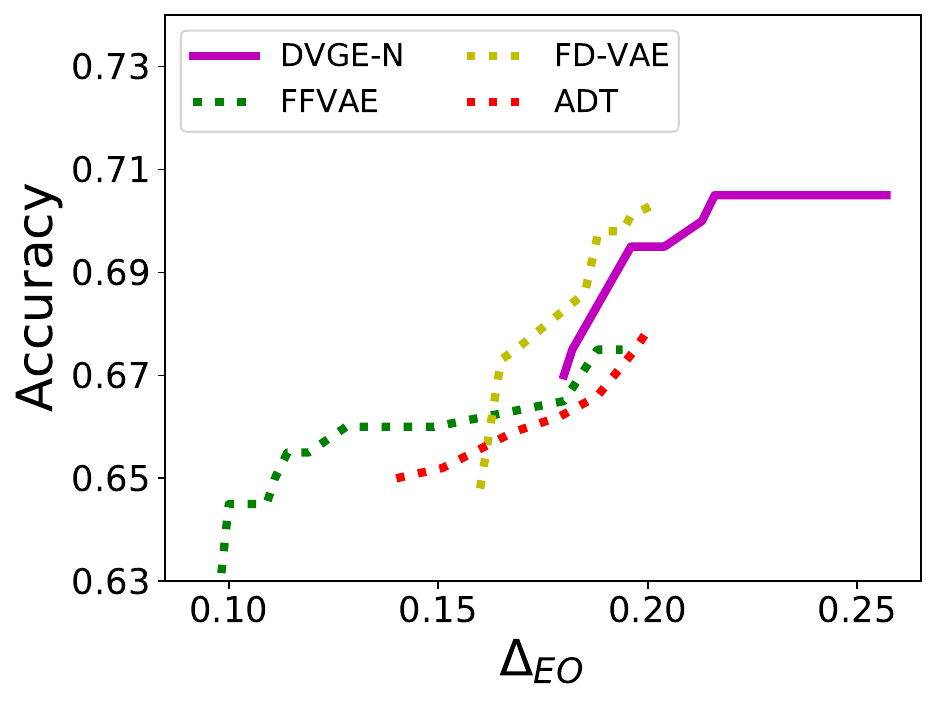}
    }
    % \vspace{-9pt}
    \caption{Fairness-accuracy trade-off comparison results for Experiment 5: South German Credit dataset, sensitive attribute = ``age"  $\land $ ``foreign\_worker", task label = ``credit\_risk".}
    \label{experiment_5}
    % \vspace{-10pt}
\end{figure*}

% \vspace{-12pt}
\subsection{Experiment Results on South German Credit}
% \vspace{-5pt}

On South German Credit, we choose two different combinations of sensitive attributes for the experiments on the structured dataset. DVGE uses the same latent code encoder for the following two different experiments.

% \vspace{-15pt}
\subsubsection{Experiment 4}
\label{sec.experiment_4}

We select ``age" as the sensitive attribute and the downstream task is to predict the label of ``credit\_risk" in this experiment. $\Delta_{DP}$ of a perfect classifier in this task would be 0.188. This experiment is designed for testing our framework when dealing with single sensitive attributes.

The experiment results are demonstrated in Figure~\ref{experiment_4}. As we can observe, when the fairness metric is $\Delta_{DP}$, both our framework and the baselines can largely reduce the unfairness of the downstream task model, but our framework achieves much higher accuracy than FFVAE and ADT. When we measure with $\Delta_{EO}$, FFVAE and ADT achieve lower values of $\Delta_{EO}$, but their downstream task accuracy is still lower than our framework. And our framework performs on par with or slightly better than FD-VAE.

% The experiment results indicate that our framework effectively reduces the loss of useful information for downstream tasks.

% \vspace{-15pt}
\subsubsection{Experiment 5}

In this experiment, we evaluate our framework when debiasing in the setting of multiple sensitive attributes in structured dataset. We consider the conjunction of ``age" and ``foreign\_worker" as sensitive attributes. The downstream task is still to predict the label of ``credit\_risk".

As we can observe in Figure~\ref{experiment_5}, the Pareto fronts suggest similar experiment results as those in the setting of a single sensitive attribute in Section~\ref{sec.experiment_4}. When the extent of fairness is measured by $\Delta_{DP}$, our framework outperforms the baselines by a large margin. When the extent of fairness is measured by $\Delta_{EO}$, the fairness-accuracy trade-off of our framework is still comparable to that of the baselines.

\begin{table*}[!t]
\centering
\caption{Debiasing performance of DVGE in the setting of single sensitive attribute}
% \vspace{-9pt}
\label{ablation_1}
\begin{adjustbox}{max width=\textwidth}
\begin{tabular}{c|c|c|cccccccccc}
\hline
\multirow{2}{*}{Encoder} & \multirow{2}{*}{No removal} & \multirow{2}{*}{\makecell{Sens. dim.\\removed}} & \multicolumn{10}{c}{DVGE with $\eta_{1}$}                                                     \\ \cline{4-13} 
                         &                             &                           & 0.1   & 0.2   & 0.3   & 0.4   & 0.5   & 0.6   & 0.7   & 0.8   & 0.9   & 1.0   \\ \hline
Disentangled                & 0.798                       & 0.736                     & 0.767 & 0.735 & 0.706 & 0.682 & 0.675 & 0.661 & 0.655 & 0.658 & 0.650 & 0.648 \\
Non-disentangled               & 0.804                       & 0.746                     & 0.769 & 0.733 & 0.705 & 0.692 & 0.686 & 0.682 & 0.682 & 0.674 & 0.671 & 0.668 \\ \hline
\end{tabular}
\end{adjustbox}
% \vspace{-5pt}
\end{table*}

\begin{table*}[!t]
\centering
\caption{Debiasing performance of DVGE in the setting of multiple sensitive attributes}
% \vspace{-9pt}
\label{ablation_2}
\begin{adjustbox}{max width=\textwidth}
\begin{tabular}{c|c|c|cccccccccc}
\hline
\multirow{2}{*}{Encoder} & \multirow{2}{*}{No removal} & \multirow{2}{*}{\makecell{Sens. dim.\\removed}} & \multicolumn{10}{c}{DVGE with $\eta_{1}$}                                                     \\ \cline{4-13} 
                         &                             &                           & 0.1   & 0.2   & 0.3   & 0.4   & 0.5   & 0.6   & 0.7   & 0.8   & 0.9   & 1.0   \\ \hline
Disentangled                & 0.752                       & 0.690                     & 0.732 & 0.707 & 0.680 & 0.661 & 0.644 & 0.638 & 0.637 & 0.633 & 0.633 & 0.631 \\
Non-disentangled               & 0.757                       & 0.704                     & 0.736 & 0.709 & 0.683 & 0.664 & 0.657 & 0.653 & 0.653 & 0.651 & 0.653 & 0.649 \\ \hline
\end{tabular}
\end{adjustbox}
% \vspace{-12pt}
\end{table*}

\subsection{Ablation}
\label{ablation}
% \vspace{-7pt}

To further evaluate the coverage on sensitive information in our framework, we conduct ablation experiments on CelebA~\cite{liu15faceattributes}. Specifically, we use the latent code perturbed by our framework to retrain sensitive classifiers. We vary the hyperparameter $\eta_{1}$ (sensitive focus) while setting $\eta_{2}=0$, and observe the highest accuracy that the retrained sensitive classifiers can achieve. Here, we use the highest accuracy of the retrained sensitive classifiers to indicate the coverage on sensitive information. The rationale of this measurement is that better coverage leads to less sensitive information in the perturbed latent code, and further the sensitive classifiers retrained with it are less accurate. \textbf{In turn, lower accuracy of the retrained sensitive classifiers indicates better coverage on sensitive information.} For comparison, we retrain sensitive classifiers using the latent code with sensitive dimensions removed~\cite{creager19flexibly} and the latent code without removal, respectively. The encoders we use here are a disentangled VAE (FactorVAE~\cite{kim18disentangling}) and a non-disentangled VAE (VanillaVAE~\cite{kingma13auto}).

First, we consider a single sensitive attribute ``Male". The ablation results are shown in Table~\ref{ablation_1}. As we can observe, when $\eta_{1}$ increases for DVGE, the highest accuracy of the retrained sensitive classifier decreases accordingly. Furthermore, when $\eta_{1}$ increases to only 0.2, DVGE achieves better coverage on sensitive information than the approach based on removing sensitive dimensions. Second, we consider two sensitive attributes, ``Male" and ``Young". The results are in Table~\ref{ablation_2}. As we can see, when $\eta_1$ increases to only 0.3, the highest accuracy of the retrained sensitive classifier with DVGE is lower than that with the approach based on removing sensitive dimensions. The ablation results demonstrate that the sensitive focus in DVGE effectively covers sensitive information.

\subsection{Discussions}

First, from the experiments above, we can observe that DVGE overall achieves better fairness-accuracy trade-off than the baselines. Second, the ablation study shows that the sensitive focus in our framework effectively covers sensitive information in the latent code. Third, we can also observe that DVGE-D generally performs better than DVGE-N from all the experiments above.

% \vspace{-10pt}
\section{Conclusion}
\label{conclusion}
% \vspace{-8pt}

In this paper, we targeted at the fairness problem in machine learning and followed the idea of using representation learning to tackle it. To overcome the problem of downstream task accuracy degradation and the problem of insufficient coverage on sensitive information, we proposed DVGE that exploits the gradient-based explanation to obtain the model focuses for respectively predicting sensitive attributes and downstream task labels, and perturbs the latent code with the focuses for the purposes of fairness and prevention of downstream task accuracy degradation. We experimentally demonstrated that our framework achieves better fairness-accuracy trade-off and better coverage on sensitive information while not relying on complete disentanglement for debiasing.

\bibliographystyle{ACM-Reference-Format}
\bibliography{sample-base}


\appendix

\section{Previous Debiasing Approaches via Removing Sensitive Dimensions Using Disentangled Representation Learning}
\label{existing_approach}

\begin{table*}[!t]
\caption{Structure of VAE, sensitive classifier, downstream task model, and ADT for CelebA}
\label{structure_1}
\centering
\begin{adjustbox}{max width=\textwidth}
\begin{tabular}{|c|c|c|}
\hline
\textbf{VAE Encoder, ADT Feature Encoder}              & \textbf{VAE Decoder}                   & \textbf{\begin{tabular}[c]{@{}c@{}}Discriminator, Sensitive Classifier,\\Dowstream Task Model, and ADT Branches\end{tabular}} \\ \hline
Input $64\times64$ image             & Input $\in\mathbb{R}^{10}$                         & Input $\in\mathbb{R}^{10}$                                                       \\ \hline
Conv2d(3,32,4,2,1) with ReLU  & Conv2d(10,256,1) with ReLU         & Linear(10,1000) with LeakyReLU(0.2)                              \\ \hline
Conv2d(32,32,4,2,1) with ReLU & ConvTrans2d(256,64,4) with ReLU    & Linear(1000,1000) with LeakyReLU(0.2)                            \\ \hline
Conv2d(32,64,4,2,1) with ReLU & ConvTrans2d(64,64,4,2,1) with ReLU & Linear(1000,1000) with LeakyReLU(0.2)                            \\ \hline
Conv2d(64,64,4,2,1) with ReLU & ConvTrans2d(64,32,4,2,1) with ReLU & Linear(1000,1000) with LeakyReLU(0.2)                            \\ \hline
Conv2d(64,256,4,1) with ReLU  & ConvTrans2d(32,32,4,2,1) with ReLU & Linear(1000,1000) with LeakyReLU(0.2)                            \\ \hline
Conv2d(256,2*10,1)            & ConvTrans2d(32,3,4,2,1)            & Linear(1000,2)                                                   \\ \hline
\end{tabular}
\end{adjustbox}
% \vspace{-7pt}
\end{table*}

Debiasing by exploiting disentangled representation learning was first proposed by Creager et al. in FFVAE~\cite{creager19flexibly} and also used by FD-VAE~\cite{park21learning}. These approaches begin with training an encoder $f(x)$ and a decoder using disentangled representation learning methods. Then, the encoder is used to produce disentangled latent code $z=f(x)$. Next, the latent code dimensions corresponding to the sensitive attributes $z_{s}$ (also known as sensitive dimensions) are determined by calculating the correlation between each dimension of $z$ and the sensitive attributes $s$ or pre-designation. At last, these approaches use the latent code without sensitive dimensions $z\backslash z_{s}$ to train downstream task models $\hat{y}=g(z\backslash{z_{s}})$. During inference, these approaches also need to remove sensitive dimensions from the latent code before feeding the code to downstream task models. In contrast, our framework DVGE does not need to make changes to the latent code during inference.

\begin{figure}[!t]
    \centering
    \subfigure[]{
    \includegraphics[width=0.15\textwidth]{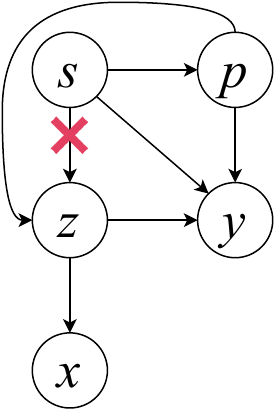}
    \label{scm_1}
    }\hspace{18pt}
    \subfigure[]{
    \includegraphics[width=0.165\textwidth]{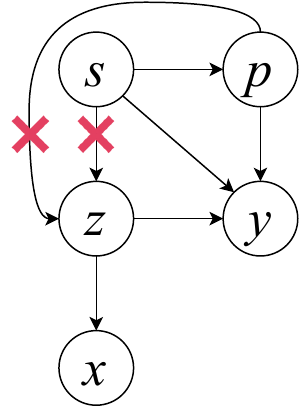}
    \label{scm_2}
    }
    % \vspace{-5pt}
    \caption{(a) Previous debiasing approaches using disentangled representation learning only break the link between the latent code $z$ and sensitive attributes $s$ in the structural causal model (SCM) when predicting downstream task attributes. (b) DVGE further breaks the link between the latent code $z$ and proxy attributes $p$.}
    % \vspace{-7pt}
\end{figure}

To further elaborate on these previous approaches, we perform a causal analysis on them by illustrating the structural causality model (SCM) of downstream tasks in Figure~\ref{scm_1}. As we can observe, because of $s\rightarrow z$ and $p\rightarrow z$, when we exploit the latent code $z$ to predict the label $y$, both sensitive attributes $s$ and proxy attributes $p$ (proxies for $s$) are considered as confounders that cause biased predictions. Since there is no guarantee of complete disentanglement from current disentangled representation learning on real-world data~\cite{nema21disentangling}, when previous debiasing approaches remove the dimensions correlated with sensitive attributes, the sensitive information from proxy attributes and some information from sensitive attributes is overlooked. As a result, in Figure~\ref{scm_1}, the link $p\rightarrow z$ is not disconnected, still causing biased predictions. In our framework, we target at breaking both $s\rightarrow z$ and $p\rightarrow z$.

\section{Implementation Details}
\label{implementation}

The platform for all the experiments in this paper is an Ubuntu 20.04 system equipped with Nvidia V100 GPUs. The implementation is based on PyTorch.

There are basically three steps to implement DVGE. First, we train VAEs to produce the latent code. Then, we train a sensitive classifier with the latent code. Finally, we train the downstream task model with the latent code according to our framework.

% \vspace{-5pt}

\begin{table*}[!t]
\caption{Structure of VAE, sensitive classifier, downstream task model, and ADT for South German Credit}
\label{structure_2}
\centering
\begin{adjustbox}{max width=\textwidth}
\begin{tabular}{|ccc|}
\hline
\multicolumn{1}{|c|}{\textbf{VAE Encoder, ADT Feature Encoder}} & \multicolumn{1}{c|}{\textbf{VAE Decoder}} & \textbf{\begin{tabular}[c]{@{}c@{}}Discriminator, Sensitive Classifier,\\Dowstream Task Model, and ADT Branches\end{tabular}} \\ \hline
\multicolumn{1}{|c|}{Input $\in\mathbb{R}^{20}$}       & \multicolumn{1}{c|}{Input $\in\mathbb{R}^{10}$}       & Input $\in\mathbb{R}^{10}$                                                       \\ \hline
\multicolumn{1}{|c|}{Linear(20,1000) with LeakyReLU(0.2)}  & \multicolumn{2}{|c|}{Linear(10,1000) with LeakyReLU(0.2)}                           \\ \hline
\multicolumn{3}{|c|}{Linear(1000,1000) with LeakyReLU(0.2)}                                                                                       \\ \hline
\multicolumn{3}{|c|}{Linear(1000,1000) with LeakyReLU(0.2)}                                                                                       \\ \hline
\multicolumn{3}{|c|}{Linear(1000,1000) with LeakyReLU(0.2)}                                                                                       \\ \hline
\multicolumn{3}{|c|}{Linear(1000,1000) with LeakyReLU(0.2)}                                                                                       \\ \hline
\multicolumn{2}{|c|}{Linear(1000,20)}                                          & Linear(1000,2)                                                   \\ \hline
\end{tabular}
\end{adjustbox}
% \vspace{-9pt}
\end{table*}

\subsection{For CelebA}
\label{app_1}

% The encoder consists of 6 layers. The first 5 layers are convolutional layers, while the last layer is to merge all channels into one for reparameterization. The decoder also consists of 6 layers which are all deconvolutional to restore the input image. For FactorVAE~\cite{kim18disentangling} and FFVAE~\cite{creager19flexibly}, since they use adversarial learning for the purpose of disentanglement, they are implemented with a discriminator which consists of 6 fully connected layers. Each layer has 1000 neurons. 

\begin{table*}[t]
\caption{The debiasing performance of DVGE in the setting of single sensitive attribute}
\label{ablation_3}
\begin{adjustbox}{max width=\textwidth}
\begin{tabular}{c|c|c|cccccccccc}
\hline
\multirow{2}{*}{Encoder} & \multirow{2}{*}{No Removal} & \multirow{2}{*}{\makecell{Sens. dim.\\removed}} & \multicolumn{10}{c}{DVGE with $\eta_{1}$}                                                     \\ \cline{4-13} 
                         &                             &                           & 0.1   & 0.2   & 0.3   & 0.4   & 0.5   & 0.6   & 0.7   & 0.8   & 0.9   & 1.0   \\ \hline
Disentangled                & 0.798                       & 0.718                     & 0.772 & 0.726 & 0.677 & 0.633 & 0.595 & 0.557 & 0.528 & 0.500 & 0.478 & 0.458 \\
Non-disentangled               & 0.804                       & 0.722                     & 0.770 & 0.719 & 0.667 & 0.621 & 0.581 & 0.545 & 0.513 & 0.489 & 0.465 & 0.447 \\ \hline
\end{tabular}
\end{adjustbox}
\end{table*}

\begin{table*}[t]
\caption{The debiasing performance of DVGE in the setting of multiple sensitive attributes}
\label{ablation_4}
\begin{adjustbox}{max width=\textwidth}
\begin{tabular}{c|c|c|cccccccccc}
\hline
\multirow{2}{*}{Encoder} & \multirow{2}{*}{No Removal} & \multirow{2}{*}{\makecell{Sens. dim.\\removed}} & \multicolumn{10}{c}{DVGE with $\eta_{1}$}                                                     \\ \cline{4-13} 
                         &                             &                           & 0.1   & 0.2   & 0.3   & 0.4   & 0.5   & 0.6   & 0.7   & 0.8   & 0.9   & 1.0   \\ \hline
Disentangled                & 0.752                       & 0.670                     & 0.732 & 0.697 & 0.662 & 0.627 & 0.599 & 0.572 & 0.549 & 0.530 & 0.513 & 0.496 \\
Non-disentangled               & 0.757                       & 0.677                     & 0.736 & 0.701 & 0.663 & 0.628 & 0.596 & 0.571 & 0.548 & 0.528 & 0.512 & 0.496 \\ \hline
\end{tabular}
\end{adjustbox}
\end{table*}

We resize the CelebA~\cite{liu15faceattributes} images to the size of $64\times64$. For a fair comparison, we implement the encoder of VAEs (VanillaVAE~\cite{kingma13auto}, FactorVAE~\cite{kim18disentangling}, FFVAE~\cite{creager19flexibly}, and FD-VAE~\cite{park21learning}) and the feature encoder of ADT~\cite{ganin16domain} with the same architecture. In terms of implementing VAEs, we follow Kim et al.~\cite{kim18disentangling} and Creager et al.~\cite{creager19flexibly} to use a CNN for the encoder, a Deconvolutional Neural Network for the decoder, and an MLP for the discriminator. The detailed structure information is shown in Table~\ref{structure_1}. To train VAEs, we set the learning rate to $10^{-4}$ and use Adam optimizer with $\beta_{1}=0.9$ and $\beta_{1}=0.999$. To train the discriminator, we set the learning rate to $10^{-5}$ and use the Adam optimizer with $\beta_{1}=0.5$ and $\beta_{1}=0.9$. The batch size is 64, and we update them for $10^6$ times (about 316 epochs). The input images are encoded into the latent codes with 10 dimensions. In terms of FFVAE, we designate the last one or two dimensions as sensitive dimensions. For FD-VAE, we designate the first three dimensions as downstream-task-related dimensions, the middle four dimensions as mutual-information dimensions, and the last three as sensitive dimensions. As for the hyperparameters of FD-VAE, we set them as in~\cite{park21learning}, As for the hyperparameters of other VAEs ($\gamma$ for FactorVAE, $\alpha$ and $\gamma$ for FFVAE), we sweep their values from 1.0 to 6.4.

Sensitive classifiers, downstream task models, and branches of ADT share the same structure with the discriminator for VAEs as shown in Table~\ref{structure_1}. To train ADT, sensitive classifiers, and downstream task models, we set the learning rate to $10^{-5}$ and use Adam optimizer with $\beta_{1}=0.5$ and $\beta_{1}=0.9$. We train sensitive classifiers for 120 epochs, and downstream task models and ADT for 100 epochs. We sweep $\eta_{1}$ and $\eta_{2}$ from 0.1 to 2.0. And we set $\epsilon_{i}$ to $0.1\times{|z_{i}|}$, where $i$ is the index for the latent code dimensions.

% \vspace{-5pt}

\subsection{For South German Credit}

The values of some attributes in South German Credit~\cite{groemping19south} are continuous, while others are categorical. To balance the value ranges, we normalize the attributes whose values are continuous with the maximal value, and convert the categorical attributes into $[0,1]$. In terms of implementing VAEs, we use MLPs for the encoder, the decoder, and the discriminator. The structure of the feature encoder of ADT is the same as that of VAE encoders. The detailed structure information for implementation is shown in Table~\ref{structure_2}. The training parameters for VAEs and the discriminator, and the latent code configurations are the same as in Appendix~\ref{app_1}.

For South German Credit, sensitive classifiers, downstream task models, and branches of ADT also share the same structure with the discriminator for VAEs as shown in Table~\ref{structure_2}. Their training parameters are also the same as in Appendix~\ref{app_1}.

\subsection{Gradient-based Explanation}

For getting gradient-based explanations from sensitive classifiers and downstream task models, we follow Srinivas et al.~\cite{srinivas19full} use predictions on the input to compute the loss for backpropagation, instead of the ground truth labels.

\section{More on Debiasing Ablation}
\label{more}

To further specifically evaluate how the sensitive focus influences the coverage on sensitive information in our framework, we design more ablation experiments on CelebA which are different from those in Section~\ref{ablation}.

In these experiments, we do not retrain sensitive classifiers, but instead directly test the accuracy of the sensitive classifiers which achieve the best accuracy before with the perturbed latent code by our framework. The perturbed latent code is generated with different configurations of the hyperparameter $\eta_{1}$ but without being perturbed by downstream task focus ($\eta_{2}=0$). Then we observe the accuracy that the sensitive classifiers can achieve. Here, we use the accuracy of the sensitive classifiers to indicate the coverage on sensitive information. The lower the accuracy is, the better the coverage on sensitive information is. To compare with our framework, we also test sensitive classifiers with the modified latent code by the existing approach and the latent code without modifications, respectively. The VAEs used in these experiments are a disentangled VAE (FactorVAE) and a non-disentangled VAE (VanillaVAE).

First, we test with the setting of a single sensitive attribute which is set to "Male". The experiment results are demonstrated in Table~\ref{ablation_3}. As we can observe, when we increase $\eta_{1}$ from 0.1 to 1.0, the accuracy of the sensitive classifier decreases from 0.772 to 0.458 for disentangled VAE, and from 0.770 to 0.447 for non-disentangled VAE, which suggests that the sensitive focus in our framework effectively covers the sensitive information with the setting of single sensitive attributes. In addition, when $\eta_{1}$ increases to only 0.2, our framework achieves comparable coverage on sensitive information with the approach based on removing sensitive dimensions. Second, we test with the setting of two sensitive attributes, which are set to "Male" and "Young". The ablation results are shown in Table~\ref{ablation_4}. As we can see, the results are similar to those in Table~\ref{ablation_3}. With $\eta_{1}$ increasing from 0.1 to 1.0, the accuracy of the sensitive classifier decreases accordingly for both disentangled VAE and non-disentangled VAE. And when $\eta_1$ is equal to or greater than 0.3, our framework outperforms the approach based on removing sensitive dimensions on the coverage on sensitive information. These results demonstrate that our framework has a good coverage on sensitive information with the setting of multiple sensitive attributes.

\end{document}